\documentclass[lettersize,journal]{IEEEtran}
\usepackage{amsmath,amsfonts}
\usepackage{algorithmic}
\usepackage{array}
\usepackage{textcomp}
\usepackage{stfloats}
\usepackage{url}
\usepackage{verbatim}
\usepackage{graphicx}
\usepackage{algorithm}
\usepackage{xcolor} 
\usepackage{multirow}
\usepackage{booktabs} 
\usepackage{makecell} 
\usepackage{subfigure}
\usepackage{bbding}
\usepackage{cite} 
\usepackage[bookmarksopen,bookmarksnumbered,citecolor=green,linkcolor=red,urlcolor=green]{hyperref}
\def\BibTeX{{\rm B\kern-.05em{\sc i\kern-.025em b}\kern-.08em
    T\kern-.1667em\lower.7ex\hbox{E}\kern-.125emX}}
\usepackage{balance}
\usepackage{bbm}
\begin{document}

\title{Image Re-Identification: Where Self-supervision Meets Vision-Language Learning}
\author{Bin~Wang, Yuying~Liang, Lei~Cai, Huakun~Huang, and Huanqiang~Zeng,~\IEEEmembership{Senior Member,~IEEE}
\thanks{\emph{(Corresponding author: Lei~Cai and Huakun~Huang)}.}
\thanks{B. Wang, Y. Liang, and H. Huang are with School of Computer Science and Cyber Engineering, Guangzhou University, Guangzhou 510006, China (e-mail: dvl.wangbin@e.gzhu.edu.cn; yuyingliang@e.gzhu.edu.cn; huanghuakun@gzhu.edu.cn).}
\thanks{L. Cai, and H. Q. Zeng are with School of Engineering, Huaqiao University, Quanzhou 362021, China (e-mail: lcai\_gxy@hqu.edu.cn; zeng0043@hqu.edu.cn).}
}

\markboth{Journal of \LaTeX\ Class Files,~Vol.~18, No.~9, September~2020}%
{How to Use the IEEEtran \LaTeX \ Templates}
\maketitle
\begin{abstract}
Recently, large-scale vision-language pre-trained models like CLIP have shown impressive performance in image re-identification (ReID). In this work, we explore whether self-supervision can aid in the use of CLIP for image ReID tasks. Specifically, we propose SVLL-ReID, the first attempt to integrate self-supervision and pre-trained CLIP via two training stages to facilitate the image ReID. We observe that: 1) incorporating~\emph{language self-supervision} in the first training stage can make the learnable text prompts more distinguishable, and 2) incorporating~\emph{vision self-supervision} in the second training stage can make the image features learned by the image encoder more discriminative. These observations imply that: 1) the text prompt learning in the first stage can benefit from the language self-supervision, and 2) the image feature learning in the second stage can benefit from the vision self-supervision. These benefits jointly facilitate the performance gain of the proposed SVLL-ReID. By conducting experiments on six image ReID benchmark datasets without any concrete text labels, we find that the proposed SVLL-ReID achieves the overall best performances compared with state-of-the-arts. Codes will be publicly available at~\href{https://github.com/BinWangGzhu/SVLL-ReID}{https://github.com/BinWangGzhu/SVLL-ReID}. 
\end{abstract}

\begin{IEEEkeywords}
Image Re-Identification, vision-language Learning, self-supervision
\end{IEEEkeywords}

\section{Introduction}
Image re-identification (ReID) is a fundamental computer vision task that aims to track a specific person/vehicle across different cameras. Such a task has widespread application prospects including intelligent video surveillance, intelligent transportation
systems, and more. However, image ReID is a highly challenging problem because images captured by different cameras could suffer from a variety of interference factors, such as clutter background, illumination variations, object occlusion,~\emph{etc}. To resist these undesirable interference factors, extracting highly representative and discriminative features to recognize the identity (ID) of a specific object is very crucial.~\emph{Convolutional Neural Network} (CNN), as a commonly suggested and very effective feature extraction pipeline, has dominated the field of image ReID for a long time. However, CNN is often blamed for only focusing on a small irrelevant region in the image~\cite{li2023clip}, this means that its extracted features are not robust and lack enough discriminative ability. In contrast, Vision Transformers (ViT) are capable of modeling the long-range dependency in the whole image without using any convolution and downsampling operations, thus are popular with many computer vision tasks. To our knowledge, for the image ReID task, nearly all CNN-based approaches~\cite{Yang2023Diverse,Yu2024Pedestrian,Khorramshahi_2023_CVPR}, ViT-based methods~\cite{Chai2023Dual,he2021transreid,zhu2023aaformer,zhu2022dual,wang2022pose,lidc23,10032729,9945658,10081216,10026500}, as well as their combination (\emph{e.g.},~\cite{Gao2023A}) need to be pre-trained on ImageNet, which contains images manually assigned one-hot labels from a pre-defined set. As a result, once the visual contents describing rich semantics lie outside the set, then they will be ignored completely~\cite{li2023clip}.



The recent work by Radford~\emph{et al.}~\cite{radford2021learning} is an exceptional contribution to the cross-modal learning between visual representation and high-level language description. This work presents large-scale~\emph{Contrastive Language-Image Pre-training} (CLIP), a novel vision-language pre-training model for connecting the visual representation with its corresponding language description. Built upon the great success of CLIP, Li~\emph{et al.}~\cite{li2023clip} made the first attempt to explore the potential of CLIP on image re-ID task. Their method, called CLIP-ReID, complements the lacking textual information by pre-training a set of learnable text tokens. Specifically, CLIP-ReID is built by pre-trained CLIP with two training stages that aim to constrain the image encoder by generating language descriptions from the text encoder. In the first training stage, a series of learnable text tokens were optimized, forming ambiguous text descriptions for each ID. In the second training stage, these ambiguous text descriptions and text encoder together impose constraints for optimizing the parameters in the image encoder, aiming at learning discriminative image features for the image ReID task.

In this work, we explore whether the momentum of self-supervised learning on both texts and images can carry into the large-scale vision-language pre-trained model. 
We note that it is not immediately clear the potential outcomes of combining self-supervision and large-scale vision-language pre-trained models like CLIP for image ReID.
It remains to be determined whether this integration will enhance the Re-ID performance or lead to conflicting information.

\begin{figure}[t]
\centerline{\includegraphics[width=0.9\linewidth]{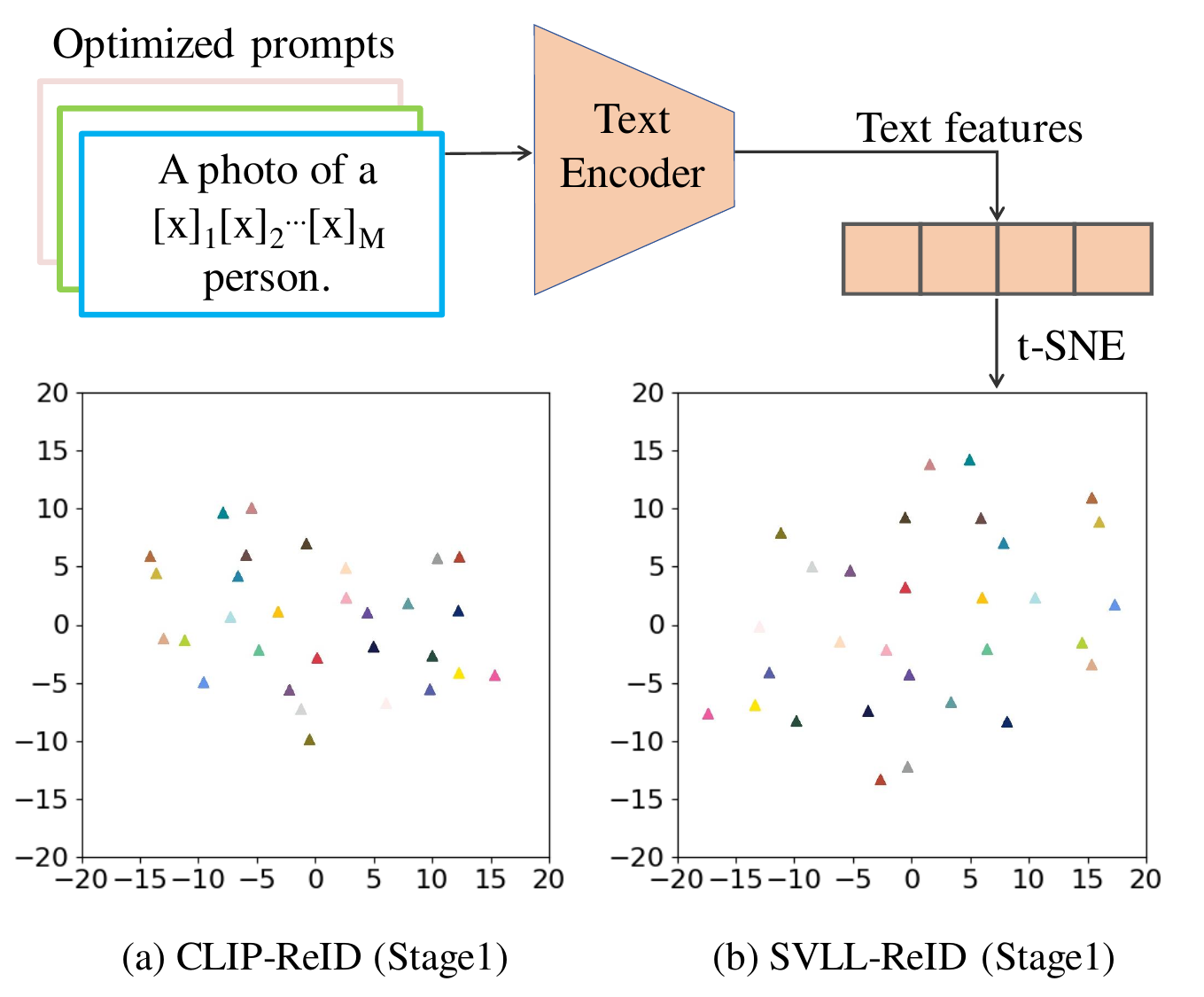}}
\vspace{-2mm}
\caption{Visualization of text feature distribution via t-SNE resulted from text encoder of both CLIP-ReID (a) and our SVLL-ReID (b) in the first stage. Different colors indicate different IDs.}
\label{fig:problem-stage1}
\end{figure}

\begin{figure}[t]
\centerline{\includegraphics[width=0.9\linewidth]{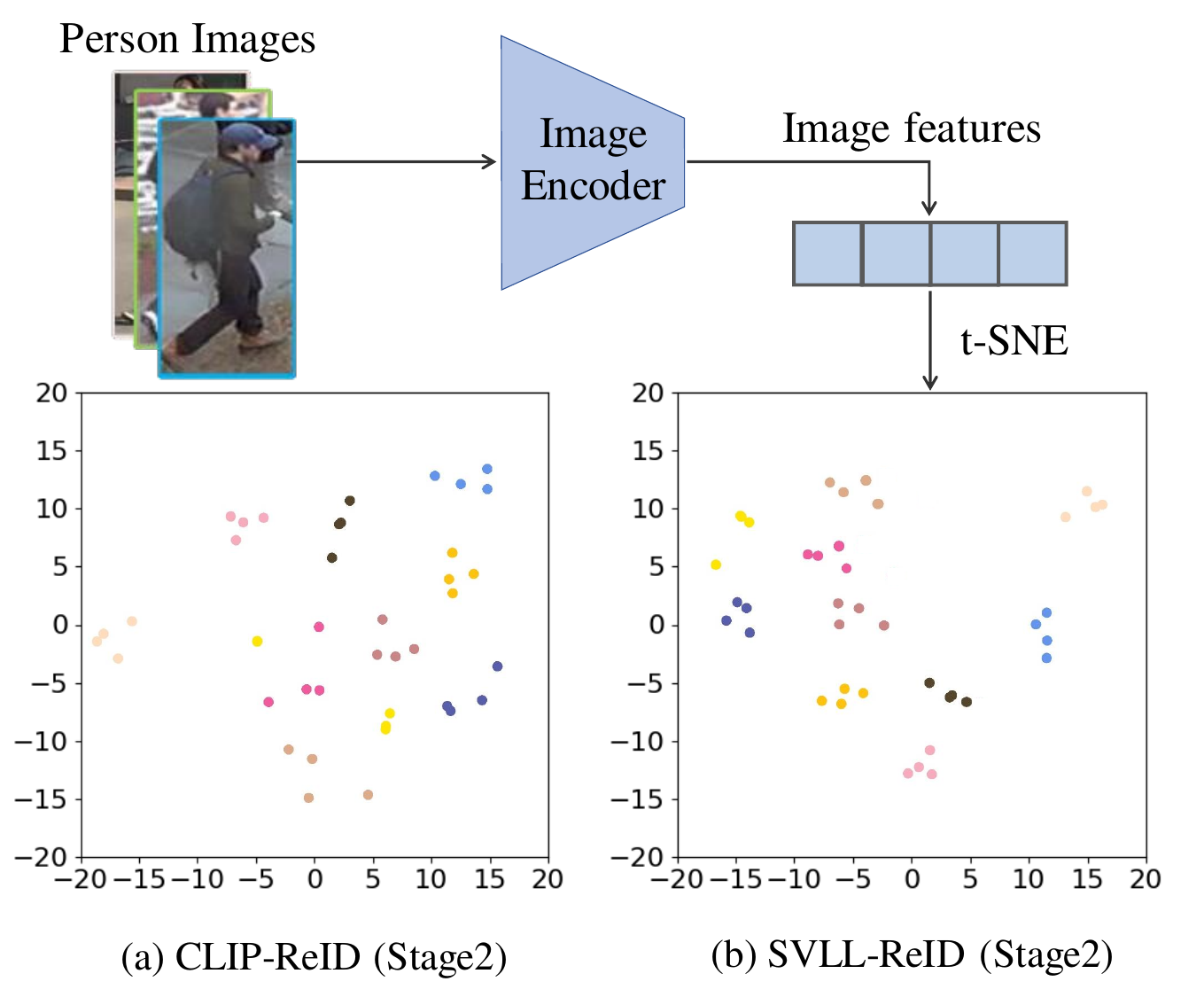}}
\vspace{-2mm}
 \caption{Visualization of image feature distribution via t-SNE from image encoder of both CLIP-ReID (a) and our SVLL-ReID (b) in the second stage. Different colors incidate different IDs.
 }
\label{fig:problem-stage2}
\end{figure}

In order to explore these questions, we propose SVLL-ReID, which is the first trial to combine self-supervision and the pre-trained CLIP via two training stages for image ReID. We empirically observe that: 1) incorporating~\emph{language self-supervision} in the first stage can make the learnable text prompts more distinguishable, as shown in Fig.~\ref{fig:problem-stage1} (a) and (b) for a comparison, and 2) incorporating~\emph{vision self-supervision} in the second stage can make the image features learned by the image encoder more discriminative, as shown in Fig.~\ref{fig:problem-stage2} (a) and (b) for a comparison. Additionally, we further validate our findings with experiments on six image Re-ID benchmarks, including person and vehicle ReID datasets without any concrete text labels. Experimental results conclusively show that the proposed SVLL-ReID outperforms multiple state-of-the-art image ReID methods, especially defeats its baseline model CLIP-ReID~\cite{li2023clip}, this is an encouraging signal for the general application of self-supervision in the context of large-scale vision-language pre-training models for image ReID. 

\begin{figure*}[t]
    \centering
    \includegraphics[width=1\textwidth]{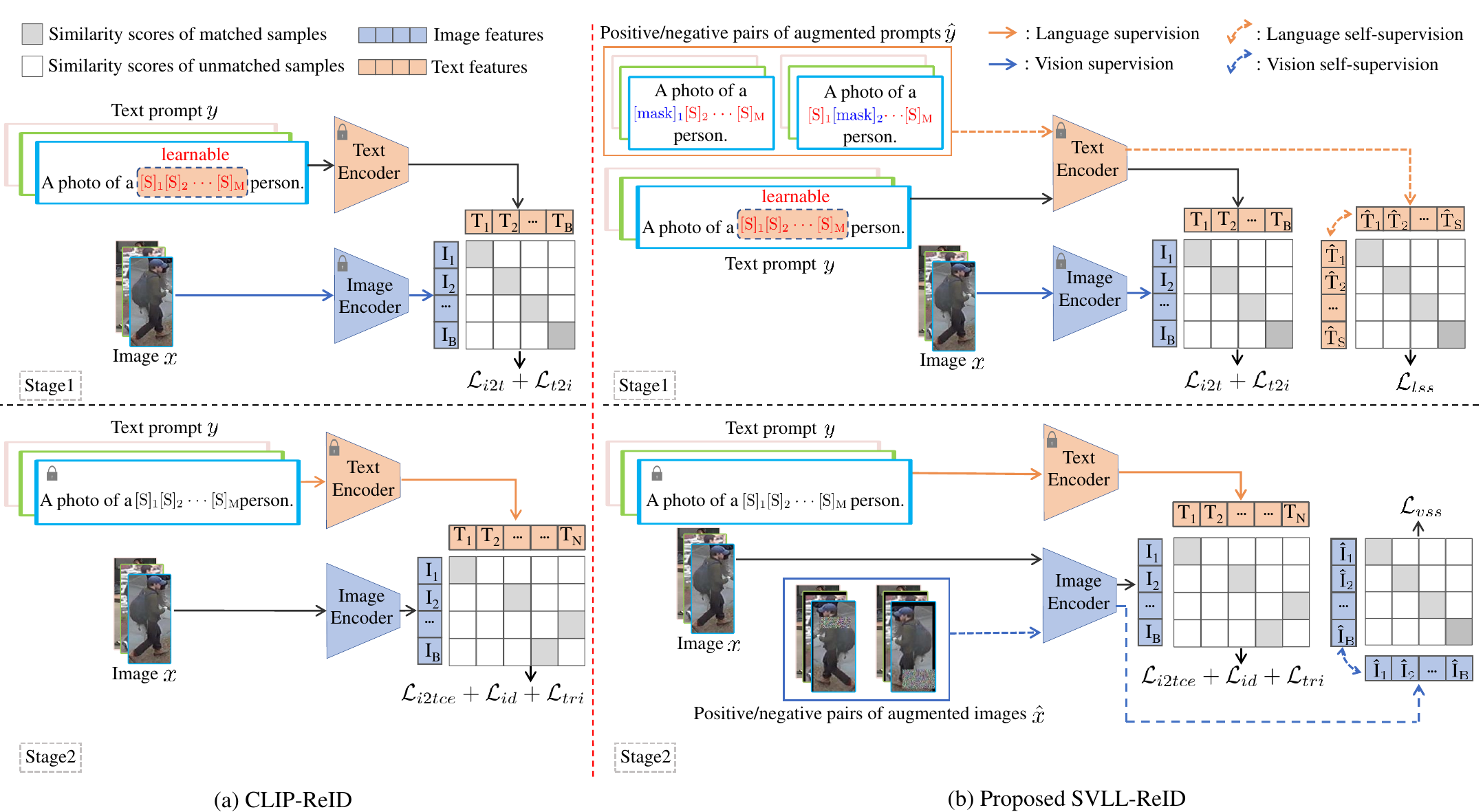}
    \vspace{-6mm}
    \caption{Comparison between CLIP-ReID~\cite{li2023clip} and the proposed SVLL-ReID. (a) is the CLIP-ReID method, which freezes the text encoder and image encoder in the first stage, and optimizes a set of learnable text tokens (\emph{i.e.}, $[S]_1$$[S]_2$$[S]_3$...$[S]_{\text{M}}$) according to~\emph{vision supervision} imposed by ReID images and the image encoder together, and then text prompts together with text encoder are to deliver~\emph{language supervision} to fine-tune the pre-trained image encoder in the second stage. (b) is the proposed SVLL-ReID, which also freezes the text encoder and image encoder in the first stage, but besides the vision supervision, the~\emph{language self-supervision} provided by augmented prompts is also introduced to help optimize $[S]_1$$[S]_2$$[S]_3$...$[S]_{\text{M}}$. Similarly, besides the language supervision, the~\emph{vision self-supervision} provided by augmented images is also introduced to help fine-tune the pre-trained image encoder in the second stage.} 
    \vspace{-2mm}
    \label{fig:framework}
\end{figure*}

\section{Methodology}
In what follows, we first briefly revisit the baseline CLIP-ReID~\cite{li2023clip} in Section~\ref{ssec:revisit}. Next, we introduce the proposed SVLL-ReID in Section~\ref{ssec:text}. The overview of SVLL-ReID compared to CLIP-ReID is shown in Fig.~\ref{fig:framework}.

\subsection{CLIP-ReID Revisit}
\label{ssec:revisit}

To be adapted to image ReID, CLIP-ReID~\cite{li2023clip} contains two training stages. The first training stage of CLIP-ReID is to generate ambiguous text descriptions for each ID. As the first stage in Fig.~\ref{fig:framework} (a) shows, CLIP-ReID replaces the original numerical labels with a set of learnable tokens for each ID. Then, these tokens are further incorporated into a text prompt $y$ such as 'A photo of a $[S]_1$$[S]_2$$[S]_3$...$[S]_M$ person'. Herein, each token $[S]_m$ ($m \in 1,...,M$) is a vector whose dimension is consistent with the word embedding, and $M$ denotes the total number of text tokens. Subsequently, the text prompt $y$ and its corresponding ReID image $x$ are separately fed into a text encoder and an image encoder to yield the text feature embedding $T$ and the image feature embedding $I$, respectively. Among them, the image feature embedding $I$ delivers vision supervision for the text prompt learning via a text-to-image contrastive loss $\mathcal{L}_{t2i}$ and an image-to-text contrastive loss $\mathcal{L}_{i2t}$. Note that the image encoder and the text encoder are frozen at this stage, and $\mathcal{L}_{t2i}$ and $\mathcal{L}_{i2t}$ are calculated to optimize the learnable tokens $[S]_1$$[S]_2$$[S]_3$...$[S]_M$, as follows:
\begin{equation}
\label{eq:li2t}
   \small \mathcal{L}_{t2i}{(T_{y_i},I_{x_i})} = -\frac{1}{|P(y_i)|}\sum_{p\in P(y_i)} \log \frac{\exp({I_{x_p}}\cdot{T_{{y_i}}})}{\sum_{k=1}^{B} \exp({I_{x_i}}\cdot{T_{{y_k}}})},
\end{equation}

\begin{equation}
\label{eq:lt2i}
   \small \mathcal{L}_{i2t}{(I_{x_i},T_{y_i})} = -\frac{1}{|P(y_i)|}\sum_{p\in P(y_i)} \log \frac{\exp({I_{x_i}}\cdot{T_{{y_p}}})}{\sum_{k=1}^{B} \exp({I_{x_k}}\cdot{T_{{y_i}}})}, 
\end{equation}
where $x_i$ and $y_i$ denote the $i$-th image and the text prompt in a batch, respectively. $P(y_i)=\{ p\in 1...B : y_p = y_i \}$ is the set of all positives with the same ID in the batch, $B$ is the batch size, and $|\cdot|$ is the cardinality of $P(y_i)$. Accordingly, the joint loss used for optimizing text tokens in the first training stage is formulated as:
\begin{equation}
\label{eq:lstage1}
\mathcal{L}_{stage1} =  \mathcal{L}_{i2t}+\mathcal{L}_{t2i}.
\end{equation}

As the second stage in Fig.~\ref{fig:framework} (a) shows, the text encoder and the text prompts (whose learnable text tokens have been optimized in the first stage) are frozen, and only the image encoder is trained with a joint loss consisting of an image-to-text cross-entropy loss $\mathcal{L}_{i2tce}$, a triplet loss $\mathcal{L}_{tri}$, and an identity loss $\mathcal{L}_{id}$ with label smoothing, aiming at making image features gather around their text prompts so that image features from different IDs become distant:
\begin{equation}
\label{eq:lstage2}
\mathcal{L}_{stage2}=\mathcal{L}_{i2tce}+\mathcal{L}_{tri}+\mathcal{L}_{id}, 
\end{equation}
where $\mathcal{L}_{tri}$ and $\mathcal{L}_{id}$ follow the definition in~\cite{luo2019bag}, whereas the image-to-text cross-entropy loss $\mathcal{L}_{i2tce}$ is defined as follows:
\begin{equation}
\label{eq:li2tce}
    \mathcal{L}_{i2tce}(I_{x_i}) = \sum_{a=1}^N -q_a\log\frac{\exp(I_{x_i}\cdot{T_{{y_a}}})}{\sum_{{k}=1}^N\exp(I_{x_i}\cdot{T_{{y_k}}})},
\end{equation}
where $q_a = (1-\epsilon) \mathbbm{1}_{[k=y]}+\epsilon/N$ is the value in the target distribution, and $\epsilon$ is a hyper-parameter which controls the degree of label smoothing. $\mathbbm{1}_{[k=y]}$ denotes an indicator function and $N$ is the total number of IDs in the dataset.

\subsection{SVLL-ReID}
\label{ssec:text}

Here, we introduce our proposed SVLL-ReID, which also contains two training stages to be adapted to image ReID.
As the first stage in Fig.~\ref{fig:framework} (b) shows, parts of tokens (\emph{i.e.}, $[S]_1$$[S]_2$$[S]_3$...$[S]_M$) in the text prompt $y$ are first randomly masked to generate new augmented prompts $\hat{y}$:
\begin{equation}
    \label{eq:masking}
     \hat{y} = Mask(y,\alpha),     
\end{equation}
where $\alpha$ represents the proportion of masked tokens in the text prompt $y$. In practice, we set the value of $\alpha$ to $0.5$. Next, two augmented prompts derived from the same text prompt form a positive pair, as the same color boxes in Fig.~\ref{fig:framework} (b) shows, while two augmented prompts from different text prompts form a negative pair, as the different color boxes in Fig.~\ref{fig:framework} (b) shows. Note that, tokens in these augmented prompts are static. After that, the positive/negative pairs are fed into the text encoder to extract the corresponding text feature embeddings. These obtained text feature embeddings deliver the~\emph{language self-supervision} to aid in the optimization of text tokens via a specifically-formulated language self-supervised loss $\mathcal{L}_{lss}$, which is devoted to pulling the text features of matched positive pairs close, while those of non-matched negative pairs far away, as follows: 
\begin{equation}
    \label{eq:tss}
    \small \mathcal{L}_{lss}(\hat{T}_{y_i}, \hat{T}_{y_j}) = -\log\frac{\exp({sim}(\hat{T}_{y_i}, \hat{T}_{y_j})/\tau)}{\sum_{k=1}^{2V} {\mathbbm{1}_{[k\neq{i}]}}\exp({sim}(\hat{T}_{y_i}, \hat{T}_{y_k})/\tau)},
\end{equation}
where $(\hat{T}_{y_i}, \hat{T}_{y_j})$ denotes the text feature embeddings resulted from a positive pair $(\hat{y}_{i},\hat{y}_{j})$, and $(\hat{T}_{y_i}, \hat{T}_{y_k})$ denotes the text feature embeddings resulted from a negative pair $(\hat{y}_{i},\hat{y}_{k})$. $V$ is the number of different prompts in a batch, $sim(u,v)$ represents the cosine similarity between the normalization embeddings $u$ and $v$, $\mathbbm{1}_{[k\neq{i}]}\in \{0,1\}$ is an indicator function whose value is equal to 1 if $k\neq{i}$, and $\tau$ is the temperature coefficient to scale the value of $sim(u,v)$.
 
By jointly minimizing the text-to-image contrastive loss $\mathcal{L}_{t2i}$ defined in Eq.(\ref{eq:li2t}), the image-to-text contrastive loss $\mathcal{L}_{i2t}$ defined in Eq.(\ref{eq:lt2i}), as well as our specifically-formulated language self-supervised loss $\mathcal{L}_{tss}$ defined in Eq.(\ref{eq:tss}), the learnable text prompts will become more distinguishable, as shown in Fig.~\ref{fig:problem-stage1} (a) and Fig.~\ref{fig:problem-stage1} (b) for a comparison. The overall formulation of $\mathcal{L}_{stage1}$ used for text prompt learning in the first stage is expressed as follows:
\begin{equation}
\label{eq:lprompts}  \mathcal{L}_{satge1}=\mathcal{L}_{t2i}+\mathcal{L}_{i2t}+\lambda_{lss}\mathcal{L}_{lss},
\end{equation} 
where $\lambda_{lss}$ is a weighting parameter that controls relative importance of $\mathcal{L}_{lss}$.





As the second stage in Fig.~\ref{fig:framework} (b) shows, parts of pixel regions in the ReID image $x$ are first randomly erased to generate new augmented image samples $\hat{x}$:
\begin{equation}
    \label{eq:erasing}
     \hat{x} = Erase(x,\beta),     
\end{equation}
where $\beta$ represents the proportion of the erased area in the image $x$. We set the value of $\beta$ to $1/3$ area size of an image in practice.
Likewise, two augmented images derived from the same ID form a positive pair, while two augmented images from different IDs form a negative pair.
These positive/negative pairs are then fed into the image encoder to yield the corresponding image feature embeddings. After that, these obtained image feature embeddings deliver the~\emph{vision self-supervision} for training the image encoder via a vision self-supervised loss $\mathcal{L}_{vss}$. Similarly, $\mathcal{L}_{vss}$ is formulated to pull the image features of matched positive pairs close, while push those of non-matched negative pairs far away:
  \begin{equation}
    \label{eq:iss}
    \small \mathcal{L}_{vss}(\hat{I}_{x_i}, \hat{I}_{x_j}) = -\log\frac{\exp({sim}(\hat{I}_{x_i}, \hat{I}_{x_j})/\tau)}{\sum_{k=1}^{2V} {\mathbbm{1}_{[k\neq{i}]}}\exp({sim}(\hat{I}_{x_i}, \hat{I}_{x_k})/\tau)},
  \end{equation}
where $(\hat{I}_{x_i}, \hat{I}_{x_j})$ denotes the image feature embeddings resulted from a positive pair $(\hat{x}_i,\hat{x}_j)$, and $(\hat{I}_{x_i}, \hat{I}_{x_k})$ represents the image feature embeddings resulted from a negative pair $(\hat{x}_i, \hat{x}_k)$. 

By jointly minimizing the image-to-text cross-entropy loss $\mathcal{L}_{i2tce}$, the triplet loss $\mathcal{L}_{tri}$, the identity loss with label smoothing $\mathcal{L}_{id}$, as well as our formulated vision self-supervised loss $\mathcal{L}_{vss}$ defined in Eq.(\ref{eq:iss}), the image features learned by the image encoder will become more discriminative, see Fig.~\ref{fig:problem-stage2} (a) and Fig.~\ref{fig:problem-stage2} (b) for a comparison. The overall formulation of $\mathcal{L}_{stage2}$ used in the second training stage is as follows:
\begin{equation}
   \label{eq:lfeature}
\mathcal{L}_{stage2}=\mathcal{L}_{i2tce}+\mathcal{L}_{id}+\mathcal{L}_{tri}+\lambda_{vss}\mathcal{L}_{vss},
\end{equation}
where $\lambda_{vss}$ is a weighting parameter for controlling the relative importance of $\mathcal{L}_{vss}$. 

\begin{table*}[!t]
	\renewcommand{\arraystretch}{0.95}
	\tabcolsep0.2cm
        \vspace{-4mm}
	\centering
	\caption{Comparison with SOTA person ReID methods on four benchmark datasets. Best results are marked in bold.}
	\begin{tabular}{ccccccccccccc}
		\Xhline{1.3pt}
		\multicolumn{1}{c}{\multirow{3}{*}{Methods}} & \multicolumn{1}{c}{Datasets} &  \multicolumn{2}{c}{Occluded-Duke~\cite{miao2019pose}}   & \multicolumn{2}{c}{DukeMTMC-ReID~\cite{zheng2017unlabeled}} & \multicolumn{2}{c}{MSMT17~\cite{wei2018person}} & 
        \multicolumn{2}{c}{Market-1501~\cite{zheng2015scalable}}  \\
		\cmidrule{2-10} & \multicolumn{1}{|c|}{References} & \multicolumn{1}{c}{mAP}    & \multicolumn{1}{c|}{Rank-1} &  \multicolumn{1}{c}{mAP}    & \multicolumn{1}{c|}{Rank-1} &  \multicolumn{1}{c}{mAP}    & \multicolumn{1}{c|}{Rank-1} &  \multicolumn{1}{c}{mAP}    & \multicolumn{1}{c}{Rank-1}  \\
		\hline
		\multirow{1}{*}{TransReID~\cite{he2021transreid}} &  \multicolumn{1}{|c|}{ICCV'21} &  \multicolumn{1}{c}{59.2}         & \multicolumn{1}{c|}{66.4}    & \multicolumn{1}{c}{82.0}         & \multicolumn{1}{c|}{90.7}  & \multicolumn{1}{c}{67.4}         & \multicolumn{1}{c|}{85.3} & \multicolumn{1}{c}{88.9}         & \multicolumn{1}{c}{95.2} \\
		\multirow{1}{*}{DCAL~\cite{zhu2022dual}} &  \multicolumn{1}{|c|}{CVPR'22} &   \multicolumn{1}{c}{-}         & \multicolumn{1}{c|}{-}      & \multicolumn{1}{c}{80.1}         & \multicolumn{1}{c|}{89.0}  & \multicolumn{1}{c}{64.0}         & \multicolumn{1}{c|}{83.1}   & \multicolumn{1}{c}{87.5}         & \multicolumn{1}{c}{94.7}   \\
		\multirow{1}{*}{PFD~\cite{wang2022pose}} &  \multicolumn{1}{|c|}{AAAI'22} &   \multicolumn{1}{c}{61.8}         & \multicolumn{1}{c|}{69.5}        &  \multicolumn{1}{c}{83.2}         & \multicolumn{1}{c|}{91.2}  &  \multicolumn{1}{c}{-}         & \multicolumn{1}{c|}{-}  &  \multicolumn{1}{c}{89.7}         & \multicolumn{1}{c}{95.5}\\
		\multirow{1}{*}{AAformer~\cite{zhu2023aaformer}} &  \multicolumn{1}{|c|}{TNNLS'23} &   \multicolumn{1}{c}{58.2}         & \multicolumn{1}{c|}{67.0}         & \multicolumn{1}{c}{80.0}         & \multicolumn{1}{c|}{90.1}  & \multicolumn{1}{c}{63.2}         & \multicolumn{1}{c|}{83.6} & \multicolumn{1}{c}{87.7}         & \multicolumn{1}{c}{95.4}   \\
		\multirow{1}{*}{DC-Foemer~\cite{lidc23}} &  \multicolumn{1}{|c|}{AAAI'23} &  \multicolumn{1}{c}{53.8}         & \multicolumn{1}{c|}{60.1}   &  \multicolumn{1}{c}{80.3}         & \multicolumn{1}{c|}{89.0} &  \multicolumn{1}{c}{70.7}         & \multicolumn{1}{c|}{86.9} &  \multicolumn{1}{c}{90.6}         & \multicolumn{1}{c}{\textbf{96.0}} \\
		\multirow{1}{*}{Kim et al~\cite{10032729}} &  \multicolumn{1}{|c|}{SPL'23} &   \multicolumn{1}{c}{62.0}     & \multicolumn{1}{c|}{70.5}            & \multicolumn{1}{c}{83.3}     & \multicolumn{1}{c|}{91.3}  & \multicolumn{1}{c}{-}     & \multicolumn{1}{c|}{-}    & \multicolumn{1}{c}{89.9}     & \multicolumn{1}{c}{95.7}\\
	\hline
		\multirow{1}{*}{CLIP-ReID~\cite{li2023clip}} &  \multicolumn{1}{|c|}{AAAI'23} &   \multicolumn{1}{c}{60.3}     & \multicolumn{1}{c|}{67.2}           & \multicolumn{1}{c}{83.1}     & \multicolumn{1}{c|}{90.8}  & \multicolumn{1}{c}{75.8}     & \multicolumn{1}{c|}{89.7}  & \multicolumn{1}{c}{90.5}     & \multicolumn{1}{c}{95.4}  \\
		\multirow{1}{*}{$\textbf{SVLL-ReID (Ours)}$} &  \multicolumn{1}{|c|}{-} & \multicolumn{1}{c}{\textbf{63.3}}  & \multicolumn{1}{c|}{\textbf{70.6}}  &  \multicolumn{1}{c}{\textbf{84.4}} & \multicolumn{1}{c|}{\textbf{91.7}} &  \multicolumn{1}{c}{\textbf{76.7}} & \multicolumn{1}{c|}{\textbf{90.1}} &  \multicolumn{1}{c}{\textbf{91.3}} & \multicolumn{1}{c}{95.9}   \\
		\Xhline{1.3pt}
	\end{tabular}

        \label{tb:person}       
        \vspace{-2mm}
\end{table*}



\section{EXPERIMENTS}
\label{sec:guidelines}

\subsection{Experimental Settings}
\subsubsection{Datasets and Evaluation Metrics}
To validate and evaluate the proposed SVLL-ReID for image Re-ID, we conducted experiments on \textbf{six} standard image datasets, including \textbf{four}~\emph{person} ReID datasets, namely Occluded-DukeMTMC~\cite{miao2019pose}, DukeMTMC-ReID~\cite{zheng2017unlabeled}, MSMT17~\cite{wei2018person}, and Market-1501~\cite{zheng2015scalable}, as well as~\textbf{two}~\emph{vehicle} ReID datasets, namely VehicleID~\cite{liu2016vehicleID} and VeRi-776~\cite{liu2016deep}. 
Note that all of person/vehicle IDs in these datasets are numeral, without concrete text labels. Regarding the evaluation metrics, we follow the general convention,~\emph{Cumulative Matching Characteristics} (CMC) at R1 and~\emph{Mean Average Precision} (mAP) were employed for quantitatively evaluating the retrieval performance.

\begin{figure}[!t]
\vspace{-1mm}
\begin{center}
\centering
   \subfigure{
        \begin{minipage}[t]{0.14\linewidth}
	\centering
	\centerline{\includegraphics[width=0.5in]{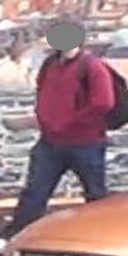}}
	\vspace{4pt}
	\centerline{\includegraphics[width=0.5in]{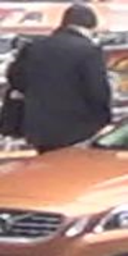}}{\scriptsize{(a)}}
			\end{minipage}%
		}%
   \subfigure{
	\begin{minipage}[t]{0.14\linewidth}
	\centering
	\centerline{\includegraphics[width=0.5in]{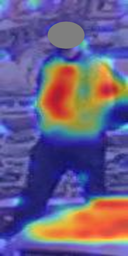}}
	\vspace{4pt}
	\centerline{\includegraphics[width=0.5in]{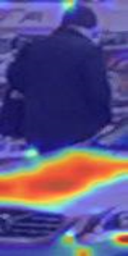}}{\scriptsize{(b)}}
		\end{minipage}%
		}%
    \subfigure{
	\begin{minipage}[t]{0.14\linewidth}
	\centering
	\centerline{\includegraphics[width=0.5in]{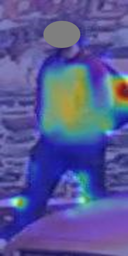}}
	\vspace{4pt}
	\centerline{\includegraphics[width=0.5in]{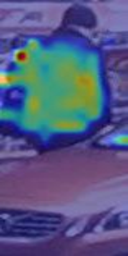}}{\scriptsize{(c)}}
		\end{minipage}%
    }%
    \subfigure{
	\begin{minipage}[t]{0.18\linewidth}
	\centering
	\centerline{\includegraphics[width=0.5in]{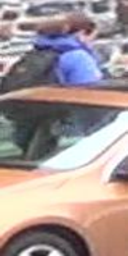}}
	\vspace{4pt}
	\centerline{\includegraphics[width=0.5in]{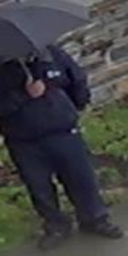}}{\scriptsize{(a)}}
		\end{minipage}%
    }%
    \subfigure{
	\begin{minipage}[t]{0.1\linewidth}
	\centering
	\centerline{\includegraphics[width=0.5in]{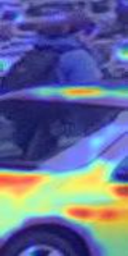}}
	\vspace{4pt}
	\centerline{\includegraphics[width=0.5in]{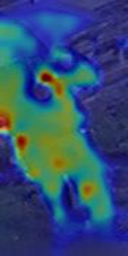}}{\scriptsize{(b)}}
		\end{minipage}%
    }%
    \subfigure{
	\begin{minipage}[t]{0.178\linewidth}
	\centering
	\centerline{\includegraphics[width=0.5in]{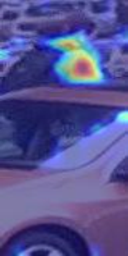}}
	\vspace{4pt}
	\centerline{\includegraphics[width=0.5in]{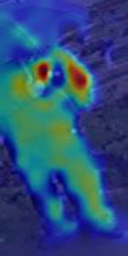}}{\scriptsize{(c)}}
		\end{minipage}%
    }%
		\centering
   \vspace{-1mm}
		\caption{Visualization of attention maps. (a) Input images, (b) CLIP-ReID, and (c) SVLL-ReID (Ours).}
		\label{fig:Hot-map}
	\end{center}
\end{figure}

\begin{table}[h]
	\renewcommand{\arraystretch}{0.95}
	\tabcolsep0.1cm
        \vspace{-2mm}
	\centering
	\caption{Comparison with SOTA vehicle ReID methods on two benchmark datasets. The best results are marked in bold.}
	\begin{tabular}{ccccccccccccc}
		\Xhline{1.3pt}
		\multicolumn{1}{c}{\multirow{3}{*}{Methods}}  & \multicolumn{1}{c}{Datasets} &\multicolumn{2}{c}{VehicleID~\cite{liu2016vehicleID}}   & \multicolumn{2}{c}{VeRi-776~\cite{liu2016deep}}  \\
		\cmidrule{2-6} &  \multicolumn{1}{|c}{References}&\multicolumn{1}{|c}{mAP}    & \multicolumn{1}{c|}{Rank-1} &  \multicolumn{1}{c}{mAP}    & \multicolumn{1}{c}{Rank-1} \\
		\hline
		\multirow{1}{*}{TransReID~\cite{he2021transreid}}  &  \multicolumn{1}{|c}{ICCV'21}         &\multicolumn{1}{|c}{90.4}         & \multicolumn{1}{c|}{85.2}  & \multicolumn{1}{c}{82.0}         & \multicolumn{1}{c}{97.1} \\
		\multirow{1}{*}{DCAL\cite{zhu2022dual}} & \multicolumn{1}{|c}{CVPR'22}         &  \multicolumn{1}{|c}{-}         & \multicolumn{1}{c|}{-}      & \multicolumn{1}{c}{80.2}         & \multicolumn{1}{c}{96.9}    \\
		\multirow{1}{*}{MART~\cite{9945658}} & \multicolumn{1}{|c}{TITS'23}         &  \multicolumn{1}{|c}{-}         & \multicolumn{1}{c|}{-}     &  \multicolumn{1}{c}{82.7}         &\multicolumn{1}{c}{\textbf{97.6}}     \\
		\multirow{1}{*}{SOFCT~\cite{10081216}}  &\multicolumn{1}{|c}{TITS'23}         &  \multicolumn{1}{|c}{89.8}         & \multicolumn{1}{c|}{84.5}  &  \multicolumn{1}{c}{80.7}         & \multicolumn{1}{c}{96.9}  \\
		\multirow{1}{*}{SSBVER ~\cite{Khorramshahi_2023_CVPR}}  &\multicolumn{1}{|c}{CVPR'23}         &  \multicolumn{1}{|c}{89.0}         & \multicolumn{1}{c|}{82.9}  &  \multicolumn{1}{c}{77.7}         & \multicolumn{1}{c}{96.0}  \\
    
		\multirow{1}{*}{GiT~\cite{10026500}}  & \multicolumn{1}{|c}{TIP'23}         & \multicolumn{1}{|c}{90.1}         & \multicolumn{1}{c|}{84.7}  &  \multicolumn{1}{c}{80.3}         & \multicolumn{1}{c}{96.9}  \\
	\hline
		\multirow{1}{*}{CLIP-ReID~\cite{li2023clip}}  & \multicolumn{1}{|c}{AAAI'23}         &  \multicolumn{1}{|c}{90.6}     & \multicolumn{1}{c|}{85.3}         & \multicolumn{1}{c}{84.5}     & \multicolumn{1}{c}{97.3}    \\
		\multirow{1}{*}{$\textbf{SVLL-ReID (Ours)}$}  &\multicolumn{1}{|c}{-}         & \multicolumn{1}{|c}{\textbf{91.3}}  & \multicolumn{1}{c|}{\textbf{86.2}}  &  \multicolumn{1}{c}{\textbf{85.1}} & \multicolumn{1}{c}{97.4}   \\
		\Xhline{1.3pt}
	\end{tabular}

        \label{tb:vehicle}       
        \vspace{-2mm}
\end{table}

\subsubsection{Implementation Details} 
The proposed SVLL-ReID was implemented with the deep learning framework PyTorch on a single NVIDIA GeForce RTX $4090$ GPU. 
For the image and text encoders used in SVLL-ReID, we adopted the pre-trained ViT-B/16 from CLIP~\cite{li2023clip} as the backbone. Like CLIP-ReID~\cite{lin2023exploring}, the proposed SVLL-ReID is also built by the vision-language pre-trained model CLIP with two training stages. In the first stage, we adopted Adam optimizer to optimize the learnable text tokens, with an initial learning rate of $3.5\times10^{-4}$ and decayed by a cosine schedule. At this stage, we randomly sampled ReID images with a batch size of $64$, and then resized them to $256 \times 128$ for person ReID and $256\times 256$ for vehicle ReID. 
$\mathcal{L}_{tss}$ in Eq.(\ref{eq:lprompts}) is set to $0.8$. In the second stage, we adopted the same warm-up strategy in CLIP-ReID~\cite{li2023clip}, which linearly grows up the learning rate from $5\times10^{-7}$ to $5\times10^{-6}$ for the initial $10$ epochs and decayed by 0.1 at the $30$-th and $50$-th epoch. Different from the sampling method in the first stage, we randomly sampled $16$ identities with $4$ images at this stage, resulting in $64$ images for each training batch. 
Likewise, $\lambda_{iss}$ in Eq.\eqref{eq:lfeature} is also set to $0.8$.


\subsection{Comparison with State-of-the-Arts }

\subsubsection{Person ReID} Table~\ref{tb:person} reports the quantitative results of the proposed SVLL-ReID and some previous state-of-the-art person ReID methods. As can be seen, the proposed SAVLL-ReID achieves the overall best performance across four person ReID datasets. This means that the proposed SVLL-ReID is more effective for the person ReID task. Additionally, from Table~\ref{tb:person}, we can also see that the proposed SVLL-ReID is superior to the baseline CLIP-ReID~\cite{lin2023exploring} as well. This demonstrates that, on one hand,~\emph{language self-supervision} in the first stage can aid in the optimization of text tokens, leading to more distinguishable text prompts for each ID, and on the other hand,~\emph{vision self-supervision} in the second stage can also assist the optimization of parameters in the image encoder, leading to more discriminative image features for person ReID. 

Additionally, another interesting observation is that, on the most challenging occluded ReID dataset named Occluded-Duke~\cite{miao2019pose}, the proposed SVLL-ReID outperforms the baseline CLIP-ReID~\cite{lin2023exploring} by a large margin, exceeding it by $3.0$ and $3.4$ in terms of mAP and Rank-1. This may be due to the fact that erasing parts of pixel regions in the image is similar to the occlusion, hence the image encoder trained with augmented images that are erased parts of pixel regions can significantly improve the retrieval ability of the person obscured by objects. To further demonstrate this, we also conduct the visualization of person images from the Occluded-Duke dataset, which applies a technique developed by Chefer~\emph{et al.}~\cite{chefer2021transformer} to display the focused areas. As can be seen from the visualization results shown in Fig.~\ref{fig:Hot-map}, CLIP-ReID~\cite{lin2023exploring} sometimes pays attention to the physical object that obscures the human body. In contrast, the proposed SVLL-ReID instead focuses more on the human body which is beneficial for the person retrieval.

\subsubsection{Vehicle ReID} In addition to person ReID, Table~\ref{tb:vehicle} shows the quantitative results of the proposed SVLL-ReID and some prior state-of-the-art vehicle ReID methods. From Table~\ref{tb:vehicle}, we observe that SVLL-ReID still achieves competitive mAP and Rank-1 performances in comparison with other competing methods. These results consistently demonstrate the effectiveness and superiority of the proposed SVLL-ReID.







\vspace{-1mm}
\section{Conclusion}
This paper investigates the effect of combining self-supervision and vision-language learning for image re-identification. In brief, we presented SVLL-ReID, the first attempt to integrate self-supervision and the vision-language pre-trained model via two training stages.
In the first training stage, we introduce~\emph{language self-supervision} to make the learnable text prompts become more distinguishable. In the second training stage, we introduce~\emph{vision self-supervision} to make the image features learned by the image encoder more discriminative. 
Experimental results on six image ReID benchmark datasets without any concrete text labels consistently show that combining self-supervision and vision-language learning indeed benefits image re-identification.




\bibliographystyle{IEEEtran}
\bibliography{reference.bib}

\end{document}